# Explaining Creative Artifacts

Lav R. Varshney [1]    Nazneen Fatema Rajani [1]    Richard Socher [1]

## Abstract

Human creativity is often described as the mental process of combining associative elements into a new form, but emerging computational creativity algorithms may not operate in this manner. Here we develop an inverse problem formulation to deconstruct the products of combinatorial and compositional creativity into associative chains as a form of post-hoc interpretation that matches the human creative process. In particular, our formulation is structured as solving a traveling salesman problem through a knowledge graph of associative elements. We demonstrate our approach using an example in explaining culinary computational creativity where there is an explicit semantic structure, and two examples in language generation where we either extract explicit concepts that map to a knowledge graph or we consider distances in a word embedding space. We close by casting the length of an optimal traveling salesman path as a measure of novelty in creativity.

*"The computer seemed to be able to see the invisible filaments that bound different ingredients together."*
*— Briscione & Parkhurst (2018)*

## 1. Introduction and Background

Due to widespread deployment of artificial intelligence (AI) technologies in settings involving people, there is growing interest in providing explanations for their results—whether to enable action, to provide a basis for evaluation, or for intrinsic reasons (Selbst & Barocas, 2018). Past work in explaining AI has focused on decisions and predictions, cf. Hind et al. (2019); Mittelstadt et al. (2019); Lage et al. (2019); here instead we consider creative ideas or artifacts. In particular, we are concerned with an inverse problem formulation and algorithms to provide post hoc human-understandable rationales for the results of computational creativity, drawing on insights from the behavioral sciences into the human process of creativity through association.

Computational creativity has advanced to the point systems are able to produce ideas and artifacts that are judged to meet standards of novelty and utility by experts in creative domains, see e.g. (Boden, 2004; 2015; Colton & Wiggins, 2012; Varshney et al., 2019; Brown et al., 2020), according to a variety of assessment criteria and theoretical frameworks (Jordanous, 2012; Colton et al., 2014; Lamb et al., 2018; Riedl, 2015; Varshney, 2019; Hashimoto et al., 2019). Although such systems may operate autonomously, they are also commonly deployed in collaboration with people to quicken innovation (Cockburn et al., 2019; Somaya & Varshney, 2020; Mehta & Dahl, 2019), often in a mixed-initiative approach (Lubart, 2005; Smith et al., 2011).

Combinatorial and compositional creativity—generation of unfamiliar combinations of familiar ideas (Boden, 2015)—is the typical kind of creativity performed by people and also pursued by computational creativity systems, whether implicitly or through explicit combining of parts. Even exceptional levels of creativity have a combinatorial character. Several algorithmic techniques now perform combinatorial creativity for different application areas, including language modeling for language (Brown et al., 2020), simulated annealing for magic tricks (Williams & McOwan, 2014; 2016), stochastic sampling + filtering (Varshney et al., 2019) as well as associative algorithms (Varshney et al., 2016) and language modeling (Lee et al., 2020) for culinary recipes, neural network approaches for music (Bretan et al., 2017) and for building materials (Ge et al., 2019), and case-based reasoning for engineering processes (Ge et al., 2018).

Though *products* of such creativity are readily interpretable by people in knowing what they are, the *pro-*

[1]Salesforce Research, Palo Alto, CA, USA. Correspondence to: Lav R. Varshney <lvarshney@salesforce.com>.

*cess* of such creativity may be inscrutable and nonintuitive (in the sense of Selbst & Barocas (2018)). Yet, users of computational creativity technologies already anthropomorphize them (Somaya & Varshney, 2018).

For human understanding, an explanatory process is often just as important as the product (Rhodes, 1961; Jordanous, 2016; Sawyer, 2019); indeed people want to understand the "theory of mind" of creators (Koelsch, 2009). This is especially the case when considering social creativity (Lee, 2014); understanding others is the most pervasive aspect of successful social interaction. In human-only creativity, the benefits of social interaction for creativity are well-known (Brass, 1995; Perry-Smith & Shalley, 2003; Perry-Smith, 2006). In this work, we develop an inverse problem formulation of going from a combinatorial artifact back to a human-like process that may have created it (even if not the underlying algorithm that actually created it).

Longstanding results in the behavioral sciences show that the way we think, recall (Howard & Kahana, 2002; Socher et al., 2009), and process information is largely through associations: an association is the connection between two or more concepts. Human creativity is often described as the mental process of combining associative elements into a new form (Barnett, 1953; Ghiselin, 1963; Bronowski, 1964; Lin & Zaltman, 1973). Mednick defined the creative process as "the forming of associative elements into new combinations which either meet specified requirements or are in some way useful" (Mednick, 1962). Although remote associations are often indicators of creativity, more nearby associations are easier for people to understand and appreciate (Grace et al., 2018). Given the human creative process is largely by association, we specifically investigate explanations using associative chains.

Note the difference from trying to explain the actual inner workings of a creativity system, such as a neural network for natural language generation (Rogers et al., 2020), which is irrelevant to the present purpose.

To find associative chains that use *nearby* associations, we develop a traveling salesman problem (TSP) formulation within knowledge graphs, where the nodes are components and edges are associations. Tours, paths, and other combinatorial structures within knowledge graphs are then possible explanations. This is the inverse problem to associative algorithms for computational creativity (Varshney et al., 2016). When relationships among components are implicit, a latent space (like word embedding in language) may make network relationships explicit. There are interesting relations to multihop reasoning over knowledge graphs (Das et al., 2017; Lin et al., 2018).

## 2. Explaining Creative Artifacts via Paths in Knowledge Graphs

Let us denote the set of components/tokens that make up the conceptual space for combinatorial or compositional creativity as the finite set $\mathcal{A} = \{a_1, a_2, \ldots, a_N\}$. The conceptual space itself is the power set of these components, $2^{\mathcal{A}}$, and a particular creative artifact, $\alpha$, is a subset of components, e.g. $\alpha = \{a_3, a_7, a_{13}, a_{47}\} \subseteq \mathcal{A}$. Although artifacts in, say, language are usually viewed syntactically as sequences of words, semantically they may be viewed as collections of concepts. There is a knowledge graph $\mathfrak{K}$ that encodes the association among these components; if relationships among components are implicit, we assume there is a similarity measure in some embedding space that yields a fully-connected graph. We assume $\mathfrak{K}$ is appropriately embedded in space so it is an undirected, weighted graph $\mathcal{K} = (\mathcal{A}, \mathcal{E})$ with nodes corresponding to the list of components and scalar-weighted edges $e \in \mathcal{E}$ that have less weight for stronger associations. If there is no association, there is infinite weight (also treated as no edge present). We assume $\mathcal{K}$ is a connected graph.

Now given a particular combinatorial artifact $\alpha_0$ to explain, we consider the subgraph of $\mathcal{K}$ restricted to the vertices $\alpha_0 \subseteq \mathcal{A}$, which we denote $\mathcal{G}$. If $\mathcal{G}$ is a connected graph, we call it $\tilde{\mathcal{G}}$. If it is not, we find a minimum-weight augmentation (Eswaran & Tarjan, 1976) using multihop paths in $\mathcal{K}$ that are collapsed into single edges to yield $\tilde{\mathcal{G}}$; this is evocative of ingredient bridging in culinary creativity (Simas et al., 2017). We further assume the graph $\tilde{\mathcal{G}}$ is *traceable* in the sense that it has at least one Hamiltonian path; this is a slightly weaker requirement than a graph being *Hamiltonian* in the sense of requiring at least one Hamiltonian cycle. Several theorems basically state that a graph is Hamiltonian (and therefore traceable) if it has enough edges (Ore, 1960). If it is not traceable, we apply a form of Hamiltonian augmentation (Di Giacomo & Liotta, 2010) by using multihop paths in the broader knowledge graph $\mathcal{K}$ to create a new graph $\tilde{\mathcal{G}}$.[1]

Now we observe that any Hamiltonian path in $\tilde{\mathcal{G}}$ (together with explanations for constituent edges from the original $\mathfrak{K}$) is an associative chain that explains the combinatorial artifact $\alpha_0$. Going from one node to the next in the Hamiltonian path is a human-interpretable associative step; traversing the whole path builds up the complete artifact. There may be many Hamiltonian paths, but as noted, nearby associations are eas-ier for people to understand (Grace et al., 2018); hence we look for a *short* (small total weight) Hamiltonian path. This view of path length may be reminiscent of regularization in other ill-posed inverse problems, but there is no tradeoff between length and fidelity here.

---

[1]If $\tilde{\mathcal{G}}$ is not traceable, an alternate approach is to use another combinatorial structure for human explanation; for example all connected graphs have minimum spanning trees which can be traversed to explain via associations.

TSP aims to find the shortest Hamiltonian path in a weighted graph, as we need. In general, any starting and ending point will do, so that choice can also be optimized. Treating an algorithm to solve TSP with this further choice as an operator, $\mathbb{TSP}(\cdot)$, the desired associative explanation is $\mathbb{TSP}(\tilde{\mathcal{G}})$. Although TSP is NP-hard, the typical artifacts we aim to explain have no more than 10 or 15 components, and so integer-linear programming may be used to find exact solutions. This is what is done for our illustrative examples in Sec. 3.

Solving the *k*-best traveling salesman problem (van der Poort et al., 1999) rather than 1-best allows several distinct associative explanations that people may find helpful. This also allows some stability of solutions (Libura et al., 1998), with explanations not changing significantly with small changes in problem parameters, which is important since unstable explanations are confusing to people.

## 3. Examples

In this section, we demonstrate our TSP-based method for explaining creative artifacts via three examples from different domains that give provide insight into different facets of the approach.

1. Culinary recipe of a new spice mixture that can be used for pastries: thyme, clove, tangerine peel oil, french lavender, and lavender flower.

2. An English sentence "After hearing the music, I woke up in the morning and opened my eyes, after which I had breakfast at the kitchen table".

3. A Hindi sentence "पुराना भारतीय सरकार अन्य अंतर्राष्ट्रीय सरकार के उल्लंघन का कारण है" (the old Indian government was the cause of other international governments unravelling).

### 3.1. Culinary Recipe (Ingredient List)

Consider a recipe (ingredient list) created by a creativity system like Chef Watson (Varshney et al., 2019), and consider the semantics of flavor pairing for providing explanation, specifically using the flavor network (Ahn et al., 2011) as knowledge graph $\mathfrak{K}$. Fig. 1 shows the 2-best $\mathbb{TSP}(\tilde{\mathcal{G}})$ for the given spice mixture. An example of an association from $\mathfrak{K}$ that is used for explanation is that "lavender flower and clove are used together since they share methyl benzoate, carvone, and linalyl acetate as flavor compounds". We observe the explanation is straightforward and insightful in this setting of explicit components and domain-specific knowledge graphs, the inverse of (Varshney et al., 2016).

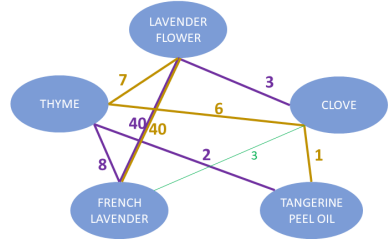

*Figure 1.* $\tilde{\mathcal{G}}$ for a novel spice mixture based on number of shared flavor compounds (more is stronger association), where two Hamiltonian paths are highlighted in purple and brown; an unused edge is in green.

### 3.2. Sentence with Explicit Knowledge Graph

The second example considers an English sentence describing a commonsense plausible routine for a human. We extracted the keywords from the text based on concepts from ConceptNet (Liu & Singh, 2004), cf. Gelfand et al. (1998). ConceptNet is a knowledge graph for representing natural language concepts with relations represented using commonsense reasoning.[2] Weights indicate the association between concepts based on their co-occurrence and the difficulty of inferring once concept given another. We consider the knowledge graph $\mathfrak{K}$ obtained by connecting relevant concepts using the commonsense relations, which is a small subgraph of the larger ConceptNet knowledge graph, as shown in Fig. 2. This is converted into $\mathcal{K}$ by ignoring edge directionality.

The 1-best $\mathbb{TSP}(\tilde{\mathcal{G}})$ serves as a plausible explanation for the extracted artifacts in our example indicated by the green path in Fig. 2. Notice that the associative path is not in the same order as the original sentence, but instead is based on strong semantic relationships. There could be multiple plausible explanation paths through different relations between concepts. Our method assumes that the generated text has a plausible commonsense explanation, but a recent line of research looks into the problem of pretrained language models generating commonsense implausible text (Lin et al., 2019).

---

[2]http://conceptnet.io/

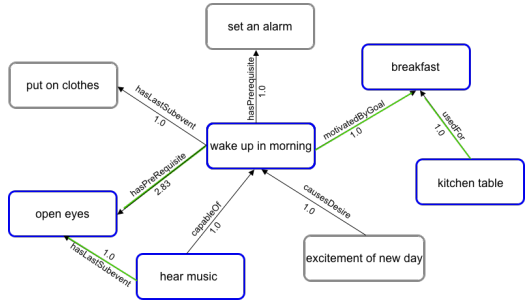

*Figure 2.* $\tilde{\mathcal{G}}$ for a novel English sentence, denoted by blue nodes and corresponding edges among them, computed using ConceptNet (ignoring directionality, larger values are stronger associations). Gray nodes are in $\mathcal{K}$ and could have been used for augmentation if needed. The path highlighted in green is a traveling salesman path.

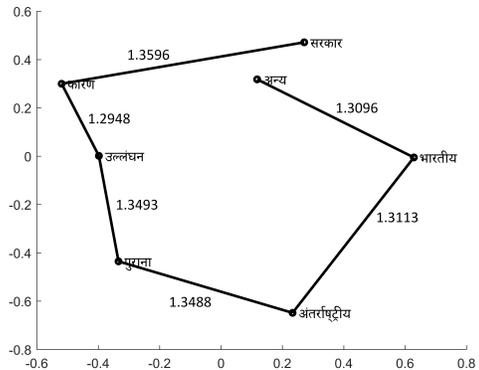

*Figure 3.* $\tilde{\mathcal{G}}$ for a novel Hindi sentence, computed using a pretrained word embedding in a two-dimensional PCA basis. Since this is a fully connected graph, we omit the unused edges in the traveling salesman path.

### 3.3. Language Sentence with Implicit Relationships

Our third example considers a sentence in a language (Hindi) less-resourced than English, where there is no commonly used equivalent to ConceptNet. There are, however, pretrained word embeddings for Hindi, which we use. In particular, we use the 300-dimensional aligned word vectors for Hindi given by Joulin et al. (2018); Bojanowski et al. (2017) to compute Euclidean distances that are in turn used to form a fully-connected graph $\mathcal{K}$ where short distances are stronger associations.

The 300-dimensional word embedding is shown in Fig. 3 by using principal components analysis to reduce dimensionality to 2 and removing the stop words. Note that this is a fully-connected graph with real-valued weights, and there is a unique traveling salesman tour found using an integer-linear program. An outer optimization is used to choose among the many options for the specific edge to break to produce the best traveling salesman path. As an example of a relationship that emerges in the explanation is between भारतीय (Indian) and अंतर्राष्ट्रीय (international). As in the English sentence example, the associative semantic path is in a different order than the sentence itself, where there is binding between concepts that are close to one another.

## 4. Traveling Salesman Path Length as a Measure of Novelty

In discussing language generation, Lin et al. (2019) argue that more distant concepts are harder to connect. Larger creative leaps are also harder for people to understand and appreciate (Grace et al., 2018): in this sense, creativity can be *too* novel. In this section, we propose that the length of a TSP path can be interpreted as a measure of novelty, since creative artifacts that connect distant concepts are very novel. Note that this is a semantic notion of novelty rather than, say, a statistical notion of novelty such as Bayesian surprise (Varshney, 2019).

**Definition 1** *Consider a creative artifact $\alpha$ comprising components $\{x_1, \ldots, x_n\}$ that has a corresponding subgraph, $\tilde{\mathcal{G}}$, of knowledge graph $\mathcal{K}$. Then the* TSP-novelty *of $\alpha$, $s_{\mathcal{K}}(\alpha)$, is defined to be $\mathbb{TSP}(\tilde{\mathcal{G}})$.*

Let us now specifically focus on the Euclidean setting as in the example from Sec. 3.3. Suppose our computational creativity algorithm selects components at random, as in stochastic sampling algorithms that, in principle, encompass all other kinds of computational creativity algorithms (Varshney, 2019). Then we have the following celebrated result in computational geometry for the Euclidean TSP problem.

**Theorem 1 (Beardwood et al. (1959))** *Let $\{X_1, \ldots, X_n\}$ be a set of i.i.d. random variables in $\mathbb{R}^d$ with bounded support. Then the length $L_n$ of the shortest TSP tour through the points $\{X_i\}$ satisfies*

$$\frac{L_n}{n^{(d-1)/d}} \to \beta_d \int_{\mathbb{R}^d} f(x)^{(d-1)/d} dx$$

*with probability 1 as $n \to \infty$, where $f(x)$ is the absolutely continuous part of the distribution of the $\{X_i\}$ and $\beta_d$ is a constant that depends on d but not on $f(x)$.*

Note that this result is for TSP tours rather than TSP paths as we consider in this work, but in fact nearly the identical result to Theorem 1 (with a slightly different constant) holds since $s_\mathcal{K}(\alpha)$ is a subadditive Euclidean functional (Steele, 1981). Of course $L_n$ is an upper bound on $s_\mathcal{K}(\alpha)$ and only slightly longer.

The intuition from this concentration of measure theorem is that, asymptotically, the choice of stochastic sampling distribution $f(x)$ in the creativity algorithm can directly control $s_\mathcal{K}(\alpha)$ in a given $d$-dimensional conceptual space. Concentrated distributions yield much less novelty than those that are disperse, and this is explicitly computable.

Moreover, note the TSP tour length $L_n$ is asymptotically, intimately tied to the Renyi entropy, $H_\gamma(f(x))$, of the sampling distribution, where for $\gamma \in (0, 1)$
$$H_\gamma(f) = \frac{1}{1-\gamma} \ln \int f^\gamma(z) dz,$$
and approaches the Shannon entropy as $\gamma \to 1$.

**Theorem 2 (Hero et al. (2002))** *Let $\{X_1, \ldots, X_n\}$ be a set of i.i.d. random variables in $\mathbb{R}^d$ with bounded support. Let $L_n$ be the length of the shortest TSP tour through the points $\{X_i\}$. Then the following estimator for Renyi entropy*
$$\hat{H}_\gamma(X) = \frac{1}{1-\gamma} \left( \ln L_n/n^\gamma - \ln \beta \right),$$
*where $\gamma = (d-1)/d$ and $\beta$ is a fixed constant independent of $f$, is an asymptotically unbiased and almost surely consistent estimator of the Renyi entropy of $f(x)$.*

That is to say, $s_\mathcal{K}(\alpha)$ is asymptotically a simple function of the Renyi-entropy of the stochastic sampling distribution, which approaches the Shannon entropy in high dimensions.

This matches with statistical measures of novelty that are also simple functions of information-theoretic quantities such as Shannon entropy and mutual information (Varshney, 2019), but here from a measure of novelty that emerges directly from explaining creative artifacts via associative chains.

## 5. Conclusion

Inspired by the psychology of human creativity, we have developed an inverse problem formulation for post hoc interpretation of creative artifacts using TSP paths in knowledge graphs, demonstrated its efficacy in a few applications, and further defined a semantic measure of novelty that is asymptotically intertwined with statistical measures of novelty.

Although our approach is built on longstanding results from the behavioral science of creativity, it remains to perform experiments with human subjects to validate our TSP-based explanations.

Going forward, it is of interest not just to produce plausible explanations of creative processes behind creative artifacts, but also to assess the *implausibility* of explanations using TSP-novelty. A large-scale study computing TSP-novelty of many artifacts made by a given AI system may measure its trustworthiness.